\pdfoutput=1
\documentclass[letterpaper, 10 pt, conference]{ieeeconf}

\IEEEoverridecommandlockouts
\usepackage{graphicx}
\usepackage{amsmath}
\usepackage{amssymb}
\usepackage{svg}
\newcommand{\etal}{\textit{et al.}}

\usepackage{hyperref}
\hypersetup{breaklinks,colorlinks}

\usepackage[capitalize]{cleveref}
\crefname{section}{Sec.}{Secs.}
\Crefname{section}{Section}{Sections}
\Crefname{table}{Table}{Tables}
\crefname{table}{Tab.}{Tabs.}

\title{\LARGE \bf
Co-Speech Gesture Synthesis using Discrete Gesture Token Learning
}

\author{Shuhong Lu$^{1}$, Youngwoo Yoon$^{2}$, and Andrew Feng$^{1}$
\thanks{$^{1}$Shuhong Lu and Andrew Feng are with Institute for Creative Technologies, University of Southern California, Los Angeles, USA
{\tt\small \{shuhongl, feng\}@usc.edu}}%
\thanks{$^{2}$Electronics and Telecommunications Research Institute, Daejeon, Republic of Korea
{\tt\small youngwoo@etri.re.kr}}%
}

\begin{document}
\maketitle
\thispagestyle{empty}
\pagestyle{empty}

\begin{abstract}
    
   Synthesizing realistic co-speech gestures is an important and yet unsolved problem for creating believable motions that can drive a humanoid robot to interact and communicate with human users. Such capability will improve the impressions of the robots by human users and will find applications in education, training, and medical services. One challenge in learning the co-speech gesture model is that there may be multiple viable gesture motions for the same speech utterance. The deterministic regression methods can not resolve the conflicting samples and may produce over-smoothed or damped motions. We proposed a two-stage model to address this uncertainty issue in gesture synthesis by modeling the gesture segments as discrete latent codes. Our method utilizes RQ-VAE in the first stage to learn a discrete codebook consisting of gesture tokens from training data. In the second stage, a two-level autoregressive transformer model is used to learn the prior distribution of residual codes conditioned on input speech context. Since the inference is formulated as token sampling, multiple gesture sequences could be generated given the same speech input using top-k sampling. The quantitative results and the user study showed the proposed method outperforms the previous methods and is able to generate realistic and diverse gesture motions. 
\end{abstract}

\section{Introduction}

Co-speech gesture synthesis enhances the realism of the robots or virtual agents by generating gesture motions that are plausible and synchronous to the corresponding speech input. It also plays an important role in making the robot move and communicate more like a real human for social interaction scenarios. Therefore realistic gesture generations would enhance the acceptances of social robots by human users and could find applications in education, training, and medical use cases. However, generating gestures that are both realistic and semantically appropriate for the input speech is a challenging and unresolved problem.

One of the challenges for generating co-speech gestures is that motion synthesis from speech is a one-to-many mapping problem, which means multiple different gestures could correspond to the same speech input. For instance, for a given speech that corresponds to a beat motion, a person could use his left, right, or both hands to perform the beat. All of them are plausible and will be deemed appropriate motions for a human user. However, learning from such data with one-to-many mapping requires more careful consideration for the model architectures as a deterministic regression will not likely learn either of the gestures but the average of them. Previous works formulated the problem as a deterministic process and therefore utilized convolutional neural network \cite{ginosar2019gestures} or recurrent neural network \cite{yoonICRA19, Yoon2020Speech} to map speech directly to gesture motions. Thus these methods tend to produce more damped gesture motions that are less interesting for human perceptions. Adding adversarial training schemes \cite{ginosar2019gestures, Yoon2020Speech} improved the results, but at the cost of a more elaborated training process that requires careful tuning. More recent works explored the probabilistic framework \cite{Alexanderson2020, Li2021} to address the non-deterministic nature of gesture synthesis and sample new gestures during inferences. This line of work builds a latent space model (\emph{e.g.}, normalizing flows) from gesture motions to learn its conditional probability distributions. Then during inference,  the gestures are randomly sampled from the latent space conditioned on speech input as the control signals. 

In this work, we tackled the issue of one-to-many mapping in gesture synthesis using discrete latent space learning. Specifically, we proposed to utilize residual quantization \cite{lee2022autoregressive} to learn discrete latent codes from gesture motions. Such discrete codes used as gesture representations allow sampling of various gesture tokens by learning an auto-regressive prior conditioned on speech context. Therefore the model is able to sample varying hand gestures associated with the same speech input with different probabilities, instead of their averaged gestures as in deterministic regressions. Residual quantization uses multiple codes to represent a gesture and reduces the reconstruction errors when using the discrete latent codes. The conditional sampling is via a two-level transformer architecture to handle the residual codes and learn the priors for gesture tokens that synthesizes diverse gesture motions. 

The proposed method is motivated by the recent success in cross-modal text-to-image synthesis \cite{dalle2021, esser2021} that utilize VQ-VAE as latent space representation for image patches and generate new images via autoregressive models to predict discrete tokens for each patch. As a high-level analogy for gesture generations, this could be considered as extracting a smaller set of gesture units from the training motions and learning the conditional probability distribution for these gesture units conditioned on speech context and previous gestures.

Our objective and subjective evaluation showed that the proposed method produces gesture motions with higher fidelity and retains the motion quality from human motion. Since the gesture synthesis process is formulated as sampling the next likely token in the codebook, multiple gesture sequences could be produced from the same speech input using top-k sampling during inference. Therefore the method does not suffer from the over-smoothed or damped gesture motions caused by the deterministic mapping of the regression models in previous methods. 

Our contributions are summarized as the following: 1) We proposed to learn discrete latent vectors as gesture motion representation through quantization to model short gesture segments as tokens. To reduce reconstruction errors and better approximate original gestures, we utilize residual quantization (RQ) during the quantization process. 2) Modeling the distribution of learned gesture tokens by using the two-level transformers to handle residual codes. With stochastic sampling, the model naturally generates multi-modal gesture motions from the same speech context. 
3) We evaluated our method in a comparison with state-of-the-art methods using both objective metrics and subjective user study. The results demonstrated that our method outperforms the previous methods in both gesture quality and diversity metrics. The user evaluation results also showed that the proposed approach can produce gesture motions with better human likeness and gesture appropriateness than the previous methods. The full source code of our implementation will be publicly available via GitHub for future research.

\section{Related Work}

\subsection{Co-Speech Gesture Generation}
The early works in co-speech gesture synthesis utilized pre-defined sets of manually created gesture units to build a gesture database. The new gestures were generated via keyword matching or prosody analysis to find the best corresponding gesture units from the database \cite{kopp2003max, kopp2009, marsella2013}. The gesture unit database can also be created automatically from speech-gesture data by segmenting and clustering gesture motions based on the similarity of motions and speech contents. During synthesis, the desired gesture property and speech attributes are used to search for the best gesture segment in the database that matches the input speech content. Our method is motivated by these ideas from early works, but instead of manually building a gesture database, we used residual quantization to implicitly learn the discrete codebook of gesture tokens. 

Recent learning-based methods train an end-to-end model from speech-gesture datasets to predict gesture motions from speech. The methods based on direct regressions find a deterministic mapping from speech to gestures \cite{kucherenko2020gesticulator, ginosar2019gestures, Yoon2020Speech, bhattacharya2021speech2affectivegestures}. Since these methods do not handle the issue of one-to-many mapping, the adversarial scheme is sometimes utilized by training the model with an additional discriminator to improve the resulting motion qualities. Recent methods further improve synthesized results by using hierarchical architecture to model multi-level skeletal poses \cite{liu2022learning} or adding semantic prompter to force semantic alignment in output gestures \cite{liang2022seeg}. 

Probabilistic frameworks were also used in the recent gesture generation works to handle the gesture ambiguities \cite{ahuja2019language2pose, Alexanderson2020, Qian2021, Li2021, bhattacharya2021text2gestures}. This type of methods learns a latent space generative model and could generate multiple gesture motions from the same speech input using conditional sampling from the latent space during inference. Similar to the previous methods, our method utilizes a latent space model and conditional sampling to handle the one-to-many problems for gesture synthesis. However, instead of learning a continuous latent space, we applied residual quantization to learn \textit{discrete} latent codes. The discrete latent codebook provides a compact representation for gesture units and the inference process is reduced to selecting the most probable latent code from the codebook. Thus by learning the conditional prior distribution over the discrete latent codes, the method naturally handles the mapping from one speech input to multiple different gestures with varying probabilities.  

\subsection{Discrete Latent Space Learning}

Vector-quantized variational autoencoder (VQ-VAE) \cite{van2017neural} learns discrete representations as a codebook from images. In the two-staged generative architecture like in Video GPT \cite{yan2021videogpt}, an autoregressive prior can then be trained to model the categorical distributions for these discrete latent codes. It was first introduced for image synthesizing or compression tasks and is able to produce sharper and higher quality image synthesis results. It was further improved in \cite{razavi2019generating} using a multi-scale hierarchical architecture to model higher-resolution images. 


One issue that often affects the reconstruction quality when training VQ-VAE is codebook collapse, which leaves a majority of the codes unused and limits the expressiveness of the model. Several methods and techniques have been proposed to prevent codebook collapse. Jukebox \cite{dhariwal2020jukebox} introduced re-initializing unused codes to a random vector to prevent dead codes during each training iteration. Video GPT \cite{yan2021videogpt} finds normalizing MSE for the reconstruction loss also mitigates codebook collapse. Hierarchical models were proposed for better codebook utilization in VQ-VAE2 \cite{razavi2019generating} by first extracting bottom and top features unconditionally to mitigate the codebook collapse. RQ-VAE~\cite{lee2022autoregressive} uses a fixed size of codebook to recursively quantize the feature map represented as a stacked map of discrete codes, which reduces the codebook size and stabilizes the codebook training. 

The discrete latent space has also been applied for text-to-image synthesis, which generates new images based on input textual description using a two-stage architecture. It first learns a discrete representation for image patches and modeling the auto-regressive priors using transformers. Learning with discrete codes is more efficient over raw pixels since the transformer may not learn the fully dependencies between pixels. The work by Esser \etal \cite{esser2021} further applied adversarial training to learn VQ-GAN that produces a perceptually rich codebook. 

In addition to image synthesis, discrete latent space model is also known to be one of the state-to-the-art methods for modeling time-series data such as audio. Jukebox \cite{dhariwal2020jukebox} utilized VQ-VAE to generate singing music. It trained multi-level networks to compress audio in different resolutions into discrete space and then used autoregressive transformers to learn the latent codes for music generation. The same idea was also adapted to generate repetitive rhythms of music by learning from extracted music loops. Multi-Instrumentalist Net \cite{su2020multi} was proposed to generate multi-instrumental music from videos, which trained VQ-VAE along with an autoregressive prior conditioned on the musician’s body key points movements. 

Compared to previous works, the proposed co-speech gesture synthesis methods utilize RQ-VAE for modeling discrete gesture tokens and a two-level RQ-Transformer architecture to model conditional priors for gesture-generating tasks. The evaluation results show its potential for retaining motion quality while allowing non-deterministic motion synthesis from the same speech input.

\section{Methods}
In this section, we briefly describe Vector Quantization and RQ-VAEs and how we applied them to the gesture synthesis problem. Then, we introduce our gesture synthesis system which consists of a text encoder, an audio encoder, and a motion generator based on transformers. 

In our training data, each speech-gesture sequence consists of speech audio, transcribed speech text, and gesture motion. Motion data is represented as a sequence of upper-body poses. For each speech-gesture sequence, we first down-sampled the motion to 20 fps and applied a sliding window of 64 frames with 10 frames step size to produce training gesture samples. Each gesture sample is represented as a tensor of size $F \times J \times R$, where $F=64$ is the sliding window size, $J$ is the number of upper-body joints (e.g.\, head, shoulder, wrist), and $R$ is the size for joint rotation representation. We also use $R=6$ as the representation for joint rotations based on previous research \cite{Zhou2019} to prevent singularities and reduce rotation approximation errors. To maintain the continuity of output gestures during synthesis, we include a 10-frame overlap between each clip for consecutive syntheses.


\begin{figure*}[h]
  \centering
  \includegraphics[width=\linewidth]{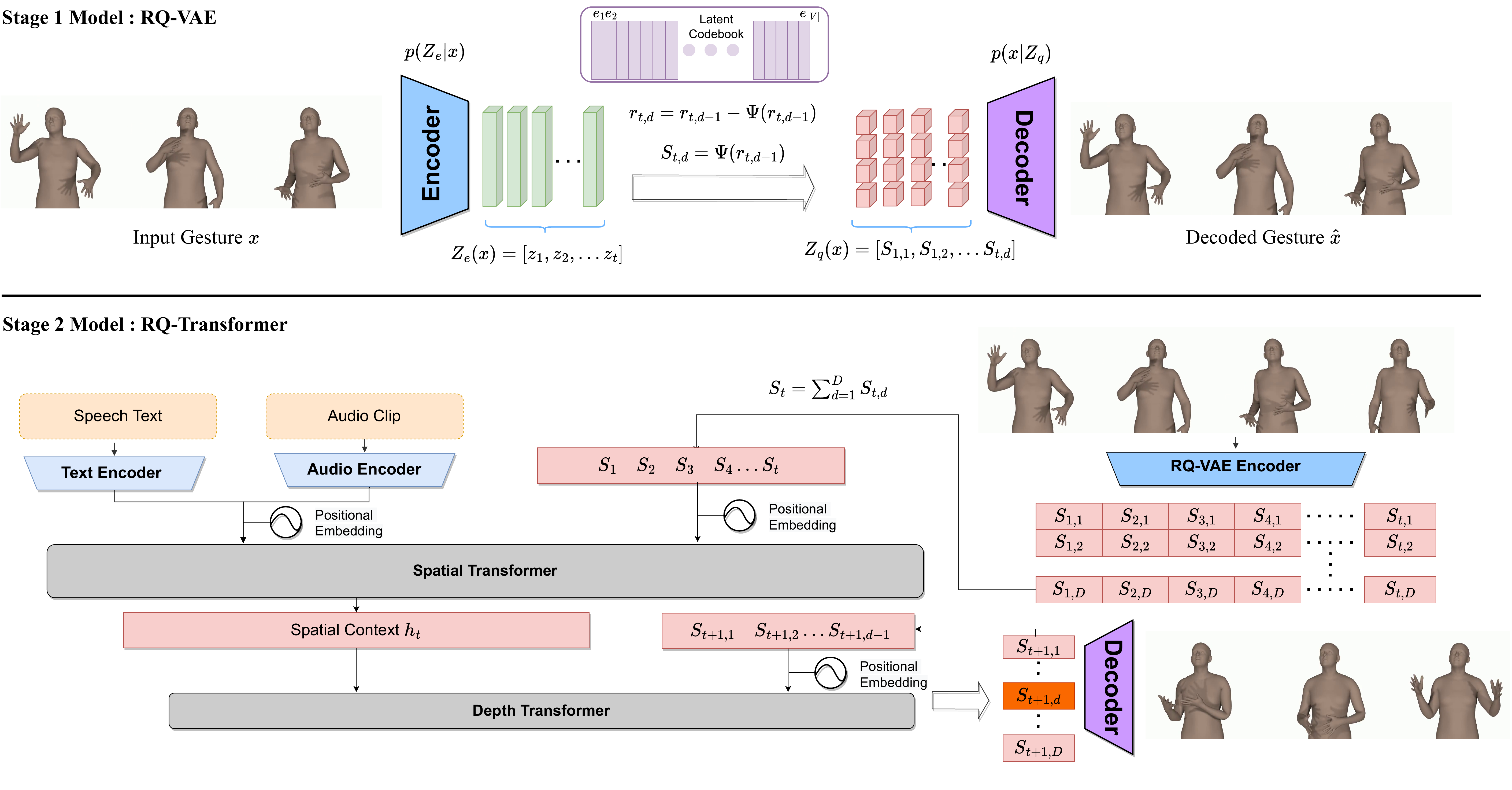}
  \caption{Overview of the two-stage model architecture for learning gesture synthesis using discrete gesture tokens. The stage 1 model learns discrete tokens via residual quantization to reconstruct input gesture motions. The stage 2 model uses the transformers to learn the autoregressive priors using speech text and audio as conditions. It first infers the spatial context $h_t$ at time step $t$ using a temporal transformer. The depth transformer then autoregressively predicts the depth token $S_{t+1, d}$ using spatial context and previously predicted depth tokens $(S_{t+1,1} \dots S_{t+1,d-1})$ at time step $t+1$. The predicted gesture tokens are then used to reconstruct the gesture motions via RQ-VAE decoder. }
  \label{fig:overview}
\end{figure*}

\subsection{Overview}


 Our method is summarized in Figure \ref{fig:overview}. Our method mainly consists of two stages that learn the gesture representations and the conditional probability distributions respectively. The first stage involves training the RQ-VAE model to learn discrete feature representation. In the second stage, we freeze the weight of RQ-VAE to treat it as a gesture encoder and use a transformer to learn the probability distribution over the discrete latent space with corresponding conditional speech features.

\subsection{Gesture Token Learning}
Discrete latent space model is adopted to extract small gesture segments as tokens from raw gesture motions (Stage 1 in Figure \ref{fig:overview}). We assume both the input and output include only gesture samples of size $F \times P$, where $F$ is the number of frames per sample and $P = J \times R$ is the pose feature size. 
The input gesture motion $x \in \mathbb{R}^{F \times P}$ is first encoded into a lower-dimensional tensor $Z_e(x) = [z_1, z_2,\ldots, z_T] $ with size $ T \times p$, where $T < F$, and $T$ indicates the number of latent vectors $z_t$ after the encoding layers. In our implementation, we use a varying number of pooling layers in the encoder so $T=F/s$ and $s$ is the temporal reduction factor. 
Then each $p$-dimensional latent vector from $Z_e(x)$ is quantized to the nearest embedding in a learnable codebook $V = \{e_{1} ,e_2,\ldots,e_{|V|}\}$, with embedding dimension $p$ and codebook size $|V|$. During the quantization stage, each feature vector from the encoder output $Z_e(x)$ is replaced by the index of the nearest vector $e_{k}$ in the codebook. 
The quantization step $\Psi$ can be summarized as:
\begin{equation}
\Psi( Z_e(x)) \ =\ e_{k} \ 
\text{where}\ k=\underset{j}{argmin} ||Z_e(x) -e_{j} ||
\end{equation}
During the reconstruction stage, the decoder takes the quantized latent vector $e$ and maps it back to the reconstructed gesture $\hat{x}$ in the original dimension. Besides the $L1$ reconstruction loss, it also includes two additional loss terms. The codebook loss helps the codebook variable training and the commitment loss is for updating encoder weights stably. The objective function for training the quantization model is then defined as the following:
\begin{equation}
L( x,\  \hat{x}) =||x-\hat{x} ||+||sg[ Z_e(x)] -e||_{2}^{2} +\beta ||sg[e] -Z_e(x) ||_{2}^{2}
\end{equation}
The operator $sg$ refers to the stop-gradient operator and $\beta$ is a hyperparameter that controls the weight of commitment loss.
The quantized result can be represented by a sequence of  the vector indices $k$ in the codebook. The decoder maps and upsamples the quantized vectors back to reconstruct the original input.
However, one big issue for training the discrete latent space model is the codebook collapse. This problem happens when only a small subset of codes are utilized in the codebook during training and will result in a latent space with less representational power. In our experiments, we found that increasing the codebook size could lead to worse reconstruction performance because of a lower code usage rate and unstable training. 

To improve code utilization, residual quantization (RQ) \cite{lee2022autoregressive} is used to represent a latent vector with multiple residual codes. It makes better usage of the codebook by recursively approximating the latent vectors with multiple codes. Specifically, for a quantization of total depth $D$, RQ recursively computes the residual code $S_{t,d}$ for the $t$-th latent vector at depth $d$ as
\begin{align}
S_{t,d}&=\Psi(r_{t,d-1}) \\
r_{t,d}&=r_{t,d-1} - \Psi( r_{t,d-1})
\end{align}
, where the 0-th residual vector is defined as the original latent vector $r_{t,0}=z_t$. $S$ is the 2D array of latent codes with size $T$ and depth $D$.
The sum of all RQ layers is denoted as $\hat{z}_t = \sum_{d=1}^D{S_{t,d}}$ which produces the final quantized vector. 
By applying the recursive quantization process, reconstruction error decreases with the increase of RQ depth. And by using a shared codebook at all depth, we are able to achieve better feature extraction under the same codebook size without increasing the number of parameters of the model, and also avoid overfitting. In our experiment, both the RQ reconstruction and inference achieve better FGD score (explained in Section \ref{sec:objective}) and lower L1 error as the number of residual layers increases. 

To further prevent codebook collapse, we also utilized several strategies proposed in previous work for learning the codebook. First, we apply exponential moving averages for the codebook learning, which places a greater weight update on the most recent codebook vectors \cite{van2017neural}. In addition, we reset the codes that are not used to random values to allow them a better chance to be utilized in the next iterations, as proposed in the Jukebox paper \cite{dhariwal2020jukebox}.
We also found that dividing the variance of the dataset when calculating MSE reconstruction loss does have a small improvement for the training \cite{yan2021videogpt}.

\subsection{Learning Conditional Priors of Gesture Token}
In the gesture synthesis stage (the second stage in Figure \ref{fig:overview}), we aim to predict the next gesture token based on previous tokens and condition vectors. The input conditions for our auto-regressive model are raw speech audio and text. We used a text encoder and an audio encoder for getting conditional feature vectors. For the speech audio features, the raw audio waveform goes through one-dimensional (1D) convolutional layers to generate a sequence of audio feature vectors. For text features, we first padded the speech text with padding tokens to make them the same length as gesture frames. 
We then used FastText~\cite{bojanowski2016enriching}, a pretrained word embedding, for speech text embedding; their weights were updated during training. And we used a temporal convolutional network (TCN)~\cite{BaiTCN2018} to process the word embeddings through a series of causal and dilated convolutions, which are shown to have advantages over traditional recurrent neural networks.
After encoding data from each modality through separate encoders, we concatenate both outputs into a single condition feature tensor $f_c$ with size $F \times (|C_{1}|+|C_{2}|)$, where $C_{1}$ and $C_{2}$ are audio and text feature vectors.

We take the feature vectors for text and audio as the condition on the transformer and do left-to-right prediction of tokens, similar to other language modeling tasks on the discrete latent codes based on transformer architecture adapted from. Specifically, we model the discrete latent variables on the residual layers as 
\begin{equation}
p( s) =\prod _{t=1}^{T}\prod _{d=1}^{D} p( S_{t,d} |S_{< t,d} S_{t,< d}, f_c)
\end{equation}
As the equation shows, we could autoregressively predict the stack of residual codes using a single transformer by expanding time and depth dimensions. To improve efficiency, we utilize the two-level architecture similar to \cite{lee2022autoregressive} by first using a Temporal Transformer (stacks of self-attention) to get context features of the tokens from previous time steps and condition vectors.
\begin{gather}
u_{t} =\textrm{PE}_T( t) +\sum _{d=1}^{D} e( S_{t-1,d} ) \ for\ t >1\\
h_{t} =\textrm{TemporalTransformer}( u_{< t}, f_c)
\end{gather}
, where PE stands for positional encodings. With the context vector on the corresponding time step, we utilize another transformer to predict code on each residual layer based on the given context vectors.
\begin{gather}
v_{t,d} =\textrm{PE}_D(d) +\sum _{d'=1}^{d-1} e( S_{t,d'} ) \ \ for\ d >1\\
v_{t,1} =\textrm{PE}_{D}(1) +h_{t} \\
p_{t,d} =\textrm{DepthTransformer}( v_{t,1}, \dots v_{t,d})
\end{gather}
Here the autoregressive steps are applied at each position $t = 1, \dotsc, T$ for the temporal transformer to get context from the previous time steps. Then the temporal context are used to predict residual layers $d = 1, \dotsc, D$ with the depth transformer and a softmax function with output probability $p_{td}$ for residual codes. In our experiment, we set temporal reduction factor $s=4$ and depth $D=4$ to allow each gesture token to represent about $0.2$ second of gesture motion. The loss is calculated using negative log-likelihood at each position of the residual layers.

\subsection{Inference}
While the model used a fixed input speech length ($F=64$ frames; 3.2 seconds), we can synthesize gesture motion for a longer speech than the fixed length by predicting $F$ frames at a time and merging them into the final gesture sequence. Each sliding window has a 10 frames overlap with the previous window and the poses are merged in the overlapped area to maintain motion continuity. Specifically, for each new synthesized gesture segment from a sliding window, its poses from the first 10 frames are linearly interpolated with the last 10 frames of the previous gesture segment before concatenation. The gesture tokens within each sliding window are predicted in a similar autoregressive manner by inferring the probability of the next token as $p_{t+1,d}$ for next time step $t+1$ and depth $d$. To allow more variety in the resulting gestures, instead of selecting the latent code with the highest probability from the codebook, we randomly sample it using the top-k probabilities where $k=10$. 


\section{Experiments}
\subsection{Datasets and Training} 
\noindent \textbf{TED Gesture} 
The dataset \cite{yoonICRA19, Yoon2020Speech} is a collection of monologue speech gestures from TED talks. The dataset consists of 1,766 videos that are processed via OpenPose \cite{OpenPose2019} to identify shots of interest. It contains $97$ hours of valid gesture motions and corresponding transcripts. For our experiments, we used the refined version \cite{feng2022tool} of the dataset by re-processing the original dataset into SMPL-X \cite{SMPL-X:2019} pose representation which gives more accurate motion. 

\noindent \textbf{Trinity} 
The dataset is a speech-gesture dataset of a single male speaker. The gestures were created via motion capture. It provides high-quality gesture motions but the dataset has a smaller size of about $5$ hours. For consistent pose representations, we retargeted the original motions to SMPL-X \cite{SMPL-X:2019} model before further data pre-processing.

We implemented our method using PyTorch framework. The codebook size was $1024$ for TED Gesture dataset and $256$ for Trinity dataset, and the temporal reduction factor was set to $s=4$ for both dataset. The RQ-VAE was trained for $500$ epoch and the RQ-Transformer was trained for $1000$ epochs. Both models were trained using AdamW optimizer with $\beta_1=0.9$ and $\beta_2=0.95$, and the learning rate was $0.0005$. We trained on a single RTX 3090 GPU for Trinity dataset using batch size $128$ and 4$\times$ RTX 8000 GPUs for TED Gesture dataset using batch size $512$.

\begin{figure}[b]
  \centering
\includegraphics[width=.95\linewidth]{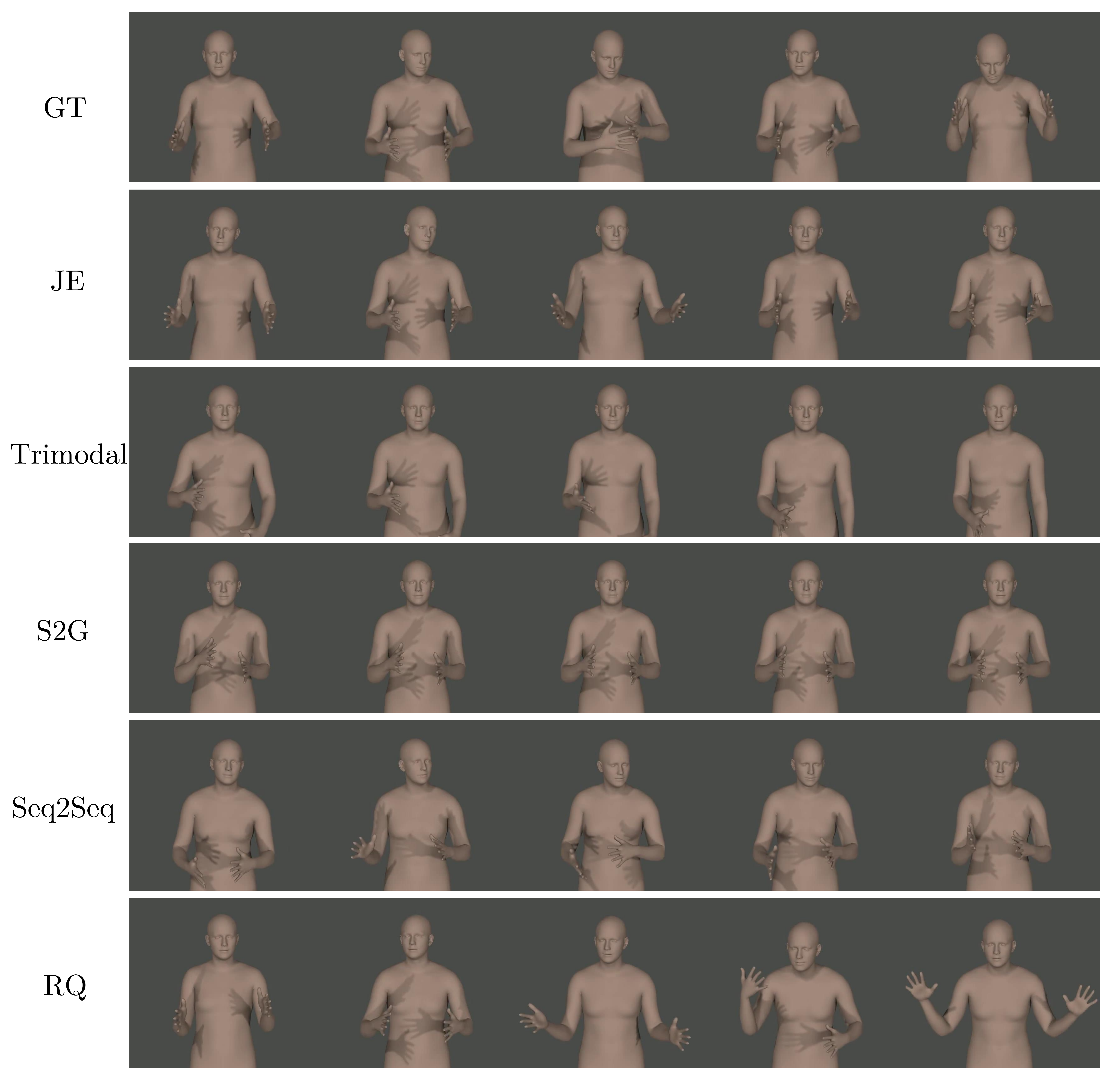}
  \caption{Samples visualization of the human motion (GT) and synthesized motions for the same speech. From top to bottom, Joint Embedding (JE), Trimodal, Speech2Gesture (S2G), Seq2Seq, and our RQ-Transformer model (RQ). Please see the supplementary video for animated visualization. }
  \label{fig:comp}
\end{figure}

\subsection{Baseline Methods}
We compared our method with five baseline methods. 1) \textbf{Seq2Seq} \cite{yoonICRA19} utilizes LSTM network with an attention mechanism to generate gestures from speech text input. 2) \textbf{Speech2Gesture} \cite{ginosar2019gestures} uses a convolutional encoder-decoder network with adversarial training to generate gestures from speech audios. 3) \textbf{Trimodal} \cite{Yoon2020Speech} uses the combination of speech audio, text, and speaker ID to synthesize gestures. 4) \textbf{Joint Embedding} \cite{ahuja2019language2pose} trains two separate encoders to map both input speech text and target motions into the same latent space before decoding them into motions. During inference, only the speech is used to obtain a latent space vector for decoding motions. 5) \textbf{VQ-Transformer} is based on our proposed model architecture without using residual quantization, and it is considered the baseline implementation.


\subsection{Objective Evaluations}\label{sec:objective}
\noindent \textbf{Evaluation Metrics}
For quantitative evaluations, we used the following metrics that were commonly used to measure the quality of synthesized gestures. \textbf{1) Fr\'echet Gesture Distance (FGD)} Similar to FID \cite{heusel2017gans} for measuring the perceptual quality of images, FGD \cite{Yoon2020Speech} measures how much the data distribution of the synthesized gestures is close to the distribution of human gestures. We followed the same encoder-decoder architectures in \cite{Yoon2020Speech} and trained the feature extractor as an autoencoder using both Trinity and TED Gesture datasets for calculating FGD.
\textbf{2) Diversity} Diversity measures how many different types of gestures are synthesized within a long sequence. We followed the formula in \cite{Li2021} by first splitting the output gestures into equal size segments and then computing the average $L1$ distances between all pairs of segments. Note that high diversity does not necessarily indicate good gestures as noisy motions could also produce high diversity results. Therefore the metric should be considered alongside FGD to ensure gesture qualities.
\textbf{3) Beat Consistency (Beat)} The metric measures the motion-audio correlations, which indicate whether the arm movements synchronize well with the beat sound in the speech audio \cite{liu2022learning}. Specifically, the shoulder, elbow, and wrist rotations are used to obtain the local optima as kinematic beats. The audio beats are then compared with kinematic beats to compute beat consistency scores as average distances between nearest pairs. 

\begingroup
\setlength{\tabcolsep}{4pt} 
\renewcommand{\arraystretch}{1} 
\begin{table}
  \caption{Objective evaluation results of the proposed method and baseline methods. Div and Beat represent diversity and beat consistency.}
  \label{tab:baseline_comp}
\centering
  \begin{tabular}{|p{2.4cm}||c|c|c||c|c|c|}
    \hline
     &
      \multicolumn{3}{c||}{TED Gesture} &
      \multicolumn{3}{c|}{Trinity }  \\
                    & FGD$\downarrow$ & Div.$\uparrow$ & Beat$\uparrow$ & FGD$\downarrow$ & Div.$\uparrow$ & Beat$\uparrow$\\
      \hline\hline
    Seq2Seq \cite{yoonICRA19}         & 8.86 & 0.91 & 0.152 & 4.99 & 1.54 & 0.238 \\
    \hline
    Speech2Gesture \cite{ginosar2019gestures} & 5.02 & 0.51 & \textbf{0.812} & 3.87 & 1.21 & 0.620 \\
    \hline
    Trimodal \cite{Yoon2020Speech}         & 1.74 & 0.65 & 0.461 & 2.86  & 1.34 & 0.652 \\
    \hline
    Joint Embedding \cite{ahuja2019language2pose} & 2.24 & 0.66 & 0.084 & 2.60  & 1.33 & 0.247 \\
    \hline
    VQ-Transformer (Ours, $D=1$)       & \underline{1.66} & \underline{1.02} & 0.793 & \underline{1.64} & \underline{1.61} &  \underline{0.733}\\
    \hline
    RQ-Transformer (Ours, $D=4$)      & \textbf{0.40} & \textbf{1.36} & \underline{0.798} & \textbf{1.52} & \textbf{1.68} & \textbf{0.745}  \\
    \hline
  \end{tabular}  
\end{table}
\endgroup

Table \ref{tab:baseline_comp} shows the quantitative results of the proposed method and baseline methods. We report FGD, Diversity, and Beat consistency measurements for both TED Gesture Dataset and Trinity Dataset. The best one is in \textbf{bold} and the second best is \underline{underlined}. For TED Gesture dataset, the RQ-Transformer performed the best in both FGD and Diversity metrics, and showed the second-highest score for beat consistency. For Trinity dataset, the RQ-Transformer obtained the best results in all metrics, which indicate synthesized gestures from the proposed method are close to human gestures and have good synchronicity with input speech. Note that the VQ-Transformer, which is the proposed method without residual quantization, is also among the top-2 methods in all metrics. Since our method encodes gestures as discrete tokens, it is less likely to produce damped motions (\emph{i.e.} for Trinity dataset, our method showed 78\% of wrist motion speed to the human motion while the other methods showed only 50--60\%) and thus has lower FGD scores. On the other hand, by using top-k sampling, the gesture generation process is not limited to deterministic mapping and is able to sample various plausible gestures based on the autoregressive prior learned from the human motion. The quantitative evaluation results from both datasets are consistent with these capabilities and show that our method is able to produce diverse gestures while retaining the gesture quality from human motion. 

\begin{table}
\caption{Experimental results on varying $D$ in Residual Quantizations (RQ).}
  \label{tab:ablation}
\centering
  \begin{tabular}{|l||c|c||c|c|c|}
    \hline
     &
      \multicolumn{2}{c||}{VAE Recon.} &
      \multicolumn{3}{c|}{Transformer Inference }  \\
                    & FGD $\downarrow$ & L1 $\downarrow$ & FGD $\downarrow$ & Diversity $\uparrow$ & Beat $\uparrow$
                    \\
      \hline\hline
    VQ ($D=1$)  & 0.85 & 0.031 & 1.64 & 1.61 & 0.733 \\
    \hline
    RQ ($D=2$)  & 0.51 & 0.027 & 1.60 & 1.61 & 0.737 \\
    \hline
    RQ ($D=3$)  & 0.28 & 0.023 & 1.57 & 1.62  & \textbf{0.751} \\
    \hline
    RQ ($D=4$)  & \textbf{0.16} & \textbf{0.022} & \textbf{1.52} & 
    \textbf{1.68}  & 0.745 \\
    \hline
  \end{tabular}  
\end{table}

\begin{table}
\caption{Experimental results on varying codebook size $N$, temporal factor $s$, and RQ layers $D$ for RQ-VAE reconstructions.}
 \label{tab:RQVAE}
\centering
\begin{tabular}{|c|c|c||c|c|}
 \hline
 $N$ & $s$ & RQ Layers & FGD & L1 Loss\\
 \hline\hline
 128 & 4 & 4 & 0.349 & 0.024 \\\hline
 256 & 4 & 4 & \textbf{0.163} & \textbf{0.022} \\\hline
 512 & 4 & 4 & 0.360 & 0.023 \\\hline
 1024 & 4 & 4 & 0.239 & \textbf{0.022} \\\hline
 256 & 2 & 2 & 0.381 & 0.730 \\\hline
 256 & 2 & 4 & 0.406 & 0.641 \\\hline
 256 & 4 & 2 & 0.515 & 0.027 \\\hline
\end{tabular}
\end{table}

We also performed an additional experiment on how RQ settings affect the gesture synthesis results by using a different number of residual layers $D$. Table \ref{tab:ablation} shows the quantitative results from the RQ-VAE and its subsequent transformers by varying $D$. Note that $D=1$ is equivalent to VQ baseline setting without residual quantization. The VAE Reconstruction part shows the gesture reconstruction results of RQ-VAE models while the Transformer Inference part shows the results from RQ-Transformer for gesture synthesis. All models were trained on Trinity dataset with codebook size 256 and temporal reduction factor $s=4$. 

For VAE reconstruction, the results indicate that with the number of RQ layers increasing, the reconstructed gestures are closer to the ground truth gestures. This is consistent with our expectations that applying the residual quantization helps reduce the reconstruction errors compared to vanilla vector quantization given the same codebook size since more codes are used to reconstruct the gesture segment.

For RQ-Transformer, the results showed that the transformer performance also increases with the corresponding VAE encoder with more residual layers, though the improvements are less significant. While RQ with $D=4$ still performs better in both FGD and diversity, it does not fully reflect the advantages of the lower reconstruction errors from VAE. The results also indicate that while increasing the RQ layers $D$ would better approximate the original data, it may also require stronger model architectures to model the conditional priors since additional residual tokens would complicate the probabilistic mapping that needs to be learned. These results validate the use of RQ over VQ for representing the gesture tokens.

Table \ref{tab:RQVAE} shows the experiment results on Trinity dataset using different codebook size $N$ and temporal reduction factor $s$ for RQ-VAE reconstructions. Intuitively, larger $N$ or smaller $s$ should result in lower reconstruction errors since the latent space model would have better capacity to represent the input data. However, in practice we found that it might result in codebook collapse and did not always produce better results. Thus these parameters are selected based on the experiment results to train the RQ-VAE and subsequent RQ-Transformer. 



\subsection{Subjective Evaluations}

\begin{figure}[t]
  \centering
  \includegraphics[width=1.0\linewidth]{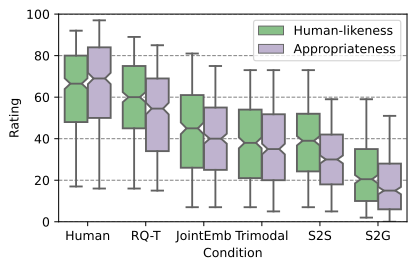} 
  \caption{Evaluation results of two subjective user studies: 1) human-likeness of motion and 2) motion appropriateness to input speech. Box plots for human motion and five gesture-generation system conditions (ordered by descending appropriateness rating on the x-axis). RQ-T represents the proposed RQ-Transformer, and the other condition names represent Joint Embedding \cite{ahuja2019language2pose}, Trimodal \cite{Yoon2020Speech}, Seq2Seq \cite{yoonICRA19}, and Speech2Gesture \cite{ginosar2019gestures}. Boxes cover 25th and 75th percentiles, and whiskers represent the 5th and 95th percentiles. Box notches represent median values.} \label{fig:userstudy_results}
\end{figure}

In addition to the objective evaluation, we conducted a subjective evaluation in that human participants rate the gesture motion videos of a virtual character. We followed the evaluation scheme introduced in GENEA Challenge 2020 \cite{kucherenko2021large} and was used in related studies \cite{yoon2021sgtoolkit, ghorbani2022zeroeggs}. The evaluation scheme consists of two studies that measure human-likeness of generated motion and appropriateness of motion to the input speech. The web interface \cite{jonell2021hemvip} presents video stimulus on a page, and a participant rates each video in a scale of 0-100. For the human-likeness study, videos were muted so that a participant can rate only the motion quality, not how appropriate to the input speech.

There were six conditions of human motion, RQ-Transformer (ours), Joint Embedding \cite{ahuja2019language2pose}, Trimodal \cite{Yoon2020Speech}, Seq2Seq \cite{yoonICRA19}, and Speech2Gesture \cite{ginosar2019gestures}. We selected 50 random speech segments from the test set and generated video stimuli corresponding to each condition for each segment using the same visualization in Figure \ref{fig:comp}. We recruited 50 valid crowd workers for each study from Prolific crowd-sourcing platform. There were 100 participants in total excluding participants who failed attention checks (36 females, 64 males; ages ranged from 19 to 70 years). Each participant rated 46 videos (8 pages x 6 videos; 2 videos were replaced by attention checks), and we collected 2300 ratings for each study. The participants were compensated 4 USD. The responses from the participants who failed the attention checks were excluded from the result analysis. In the attention-check videos, an overlay text requesting to rate the current video to a specific number appears in the middle of the video playing.

Figure \ref{fig:userstudy_results} shows the results of two studies. The human motion condition showed the highest ratings in both studies (Median$=$67 for the human-likeness study, 69 for the appropriateness study). The RQ-Transformer showed the second highest ratings for both studies (Median$=$60, 55). Following them, the other conditions showed median ratings of 45, 40 (Joint Embedding); 38, 35 (Trimodal); 39, 30 (Seq2Seq); 21, 15 (Speech2Gesture). The ratings between the conditions were significantly different except for between Trimodal and Seq2Seq in the human-likeness study. We used Wilcoxon rank-sum tests with Holm–Bonferroni corrections for multiple comparisons. The family-wise error rate was 0.05.

The subjective evaluation showed a consistent result to the objective evaluation that the proposed RQ-Transformer is able to synthesize more natural and appropriate gestures than the previous methods. In both subjective evaluation studies, RQ-Transformer was rated higher than all the other generation methods by large margins. The difference between the natural human motion and the proposed method was small in terms of human-likeness of motion while the difference was apparent in terms of speech appropriateness. Conveying semantic gestures like iconic and metaphoric gestures is a challenging problem especially for data-driven gesture generation approaches because of the scarcity of data samples mapping speech to semantic gestures.

Figure \ref{fig:comp} showed examples of the qualitative results by comparing gesture results from our method and baseline methods with natural human motions. Here the keyframes from the synthesized gestures of our method and other baselines were visualized and compared against the ground truth motions. Joint Embedding \cite{ahuja2019language2pose} and Seq2Seq \cite{yoonICRA19} methods produce damped motions that are less synchronous to input speech. For Speech2Gesture \cite{ginosar2019gestures} and Trimodal \cite{Yoon2020Speech} methods, they tend to produce synchronous gesture motions but lack motion diversity when compared to ground truth. By using discrete gesture tokens as data representation, our model is able to produce more interesting and diverse gesture motions than other baselines. Please see the supplemental video for more qualitative comparison results. 

\section{Conclusion}
In this work, we proposed to use discrete gesture tokens for learning gesture synthesis model. The method utilizes RQ-VAE for learning gesture units and a two-level autoregressive transformer for learning conditional latent code priors. By using discrete tokens as gesture representations, it reduces the problem of gesture synthesis into learning the probability distributions for categorical token indices. Such setup naturally addresses the one-to-many mapping issue in gesture synthesis by assigning different probabilities to multiple gesture tokens even if they are corresponding to the same speech. Since the method generates gestures by selecting the next likely tokens based on the learned probability distributions, it is able to produce varying gesture motions given the same speech context. Both the objective and subjective evaluation results show that the method is able to generate diverse gesture motions with adequate human-likeness.  

\textbf{Limitations and Future Works}
While our method performed relatively well in the user evaluation studies, some limitations need to be considered and addressed in future works. First, vector quantization approximates the original data and might produce lower-quality motions with less variety. Although we partially addressed this issue by using residual quantization to reduce the reconstruction errors, information would still be lost during quantization. Second, since each token contains multiple frames of gesture motions, it may limit the temporal resolution of resulting motions since gestures could only change after predicting the next token. Thus the synthesized gestures might be less aligned with input speech compared to methods that synthesize the motions frame-by-frame. Finally, while residual quantization produces better reconstruction results, it requires more burdens on learning the autoregressive priors and thus may need more powerful networks to handle the complicated relationships between many tokens when $D$ is large. 
{\small
\bibliographystyle{ieee_fullname}
\bibliography{shortstrings,bibliography}
}

\end{document}


\title{Supplementary Materials for Co-Speech Gesture Synthesis using Discrete Gesture Token Learning }

\author{First Author\\
Institution1\\
Institution1 address\\
{\tt\small firstauthor@i1.org}
\and
Second Author\\
Institution2\\
First line of institution2 address\\
{\tt\small secondauthor@i2.org}
}

\maketitle

\begin{figure}[h]
  \centering
  \includegraphics[width=0.9\linewidth]{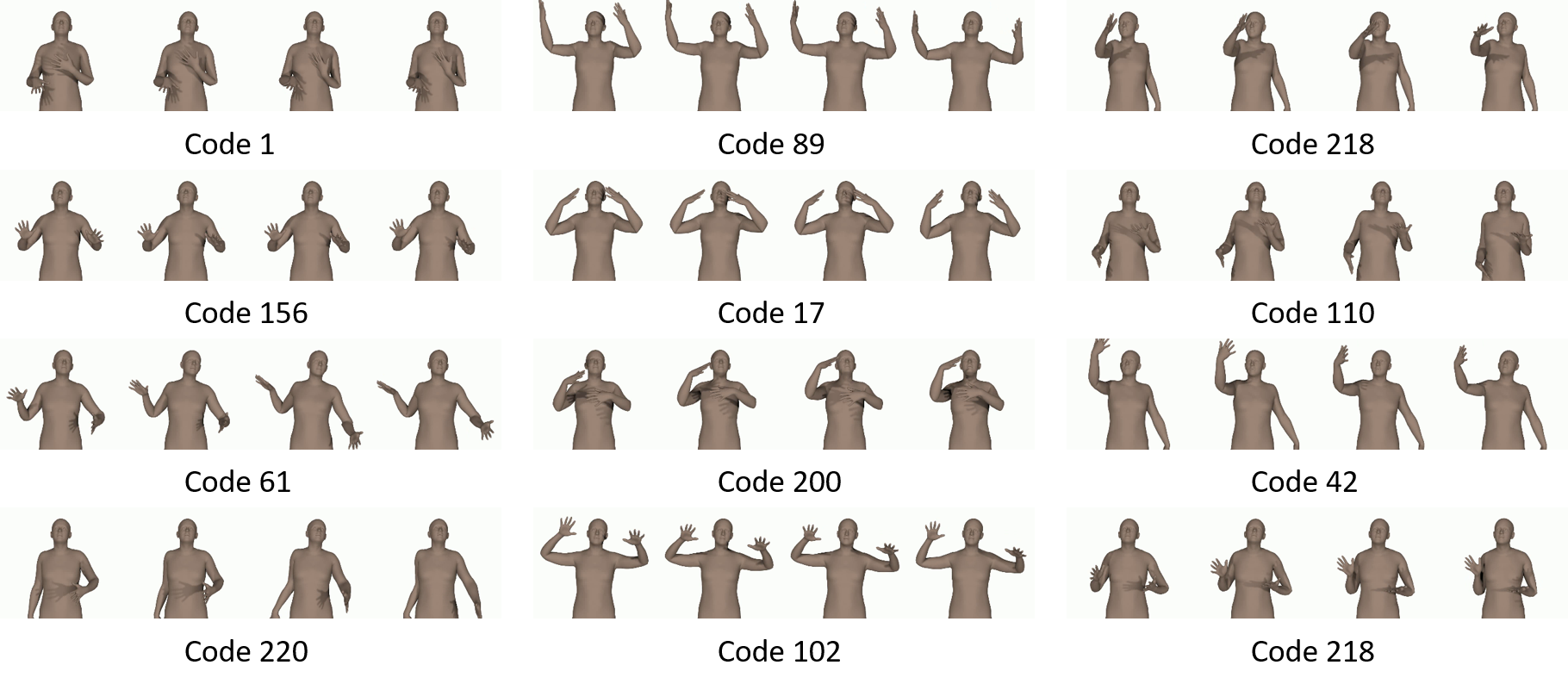}
  \caption{Visualization of exemplar discrete gesture tokens with corresponding code indices. Each token consist of 4 motion frames showing representative gesture poses. With a properly learned codebook, a variety of gesture types are compactly represented as different discrete latent codes. }
  \label{fig:vq_vis}
\end{figure}

\section{Visualization of Discrete Gesture Tokens}
To examine the learned gesture tokens visually, we decode each gesture latent vector in the codebook into a 4-frame gesture segment and render the motions. Figure \ref{fig:vq_vis} shows the visualization results from a subset of gesture codes learned from Trinity dataset with codebook size $N=256$. Note that the residual quantization (RQ) combines multiple codes together to form a more accurate reconstruction of the input gesture data. Thus even with a relatively small codebook size $N=256$, it can produce $O(N^4)$ code combinations to form different gesture segments.

\section{User Study Interface Details}

\begin{figure}[h]
  \centering
  \includegraphics[width=0.6\linewidth]{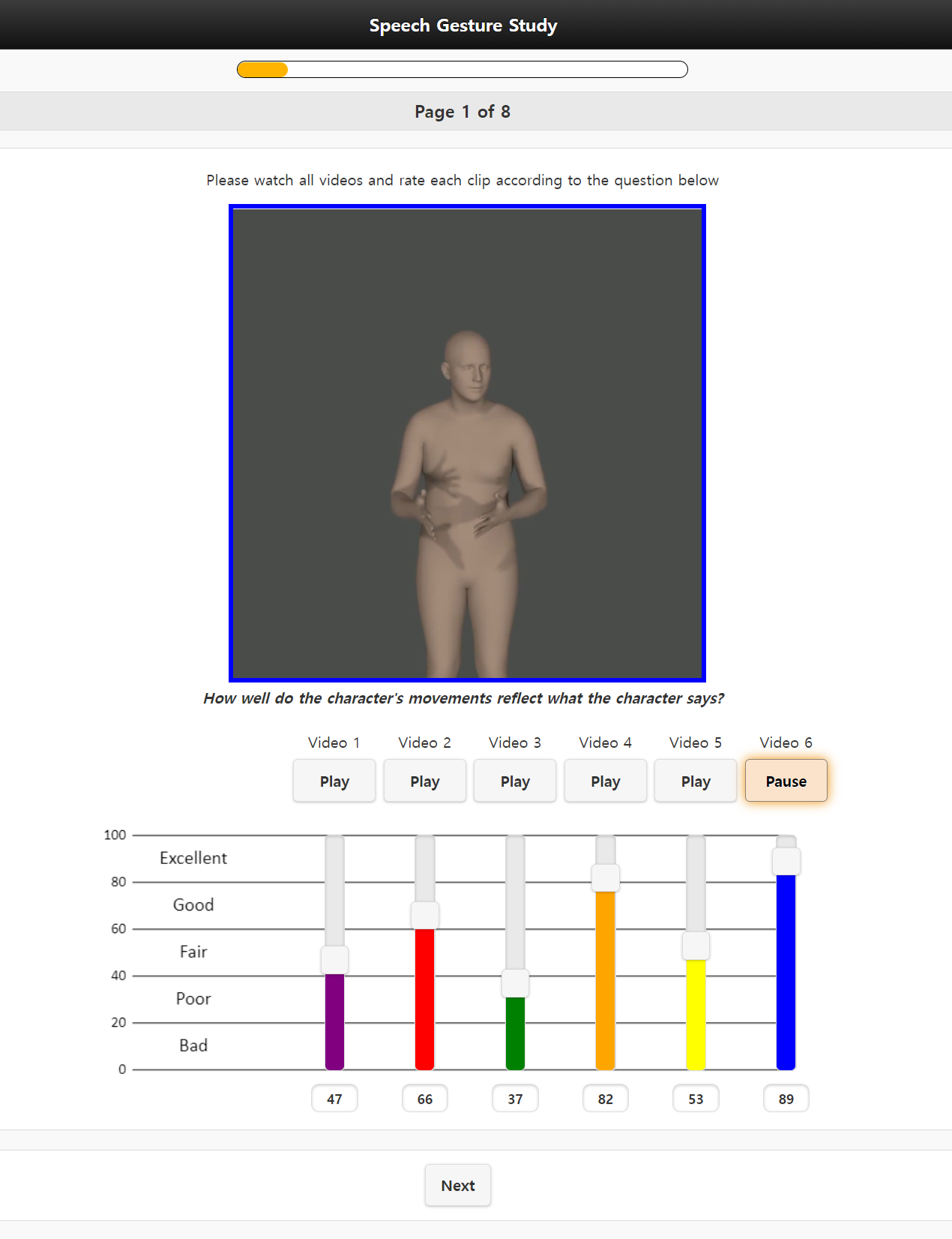}
  \caption{HEMVIP interface \cite{jonell2021hemvip} used in the user studies.}
  \label{fig:hemvip_interface}
\end{figure}

Figure \ref{fig:hemvip_interface} shows the user interface that we used in the user studies. We used the question ``How humanlike does the gesture motion appear?'' for human-likeness studies, and we used ``How well do the character's movements reflect what the character says?'' for the appropriateness studies.

\section{Additional Experiments}
We have experimented on training RQ-VAE and its corresponding transformers using different codebook size $N$ and temporal reduction factor $s$. We also investigated how varying $k$ in the top-$k$ sampling affects the resulting gestures. The Trinity dataset \cite{IVA:2018} was used for all of the experiments. 

Table \ref{tab:RQVAE} shows the results of using different codebook size $N$ and temporal reduction factor $s$. It tested how varying these parameters could affect the quality of the generated gesture clips. Inuitively, larger $N$ or smaller $s$ should result in lower reconstruction errors since the latent space model would have better capacity to represent the input data. However, in practice we found that it might result in codebook collapse and did not always produce better results. In general, increasing the number of RQ layers tend to reduce the reconstruction errors and the decoded gestures would be closer to the ground truth gestures. Based on these quantitative results, we picked the RQ-VAE model with codebook size $N=256$ and temporal factor $s=4$ for training the subsequent transformer used in the objective evaluations of the main paper. 

Table \ref{tab:RQTransformer} shows how varying the parameter $k$ in top-$k$ sampling affects the quality in the synthesized gesture clips. Instead of greedily picking the most probable tokens, the top-$k$ sampling randomly select a token from the highest $k$ probabilities during inference. From the results, we found that for all conditions, applying top-$k$ sampling usually improved the results. By using $k=50$, the model produced diverse gesture results with better quality than using smaller $k$.

\begin{table}
\begin{tabular}
{ 
  | >{\centering\arraybackslash}X
  | >{\centering\arraybackslash}X 
  | >{\centering\arraybackslash}X
  | >{\centering\arraybackslash}X
  | >{\centering\arraybackslash}X | }
 \hline
 \multicolumn{5}{|c|}{Experiments for RQ-VAE Reconstruction } \\
 \hline
 Codebook Size $N$ & Temporal Factor $s$ & Num RQ Layer & FGD& L1 Loss\\
 \hline
 128 & 4 & 4 & 0.349 & 0.024 \\
 256 & 4 & 4 & 0.163 & 0.022 \\
 512 & 4 & 4 & 0.360 & 0.023 \\
 1024 & 4 & 4 & 0.239 & 0.022 \\
 256 & 2 & 2 & 0.381 & 0.730 \\
 256 & 2 & 4 & 0.406 & 0.641 \\
 256 & 4 & 2 & 0.515 & 0.027 \\
 \hline
\end{tabular}
 \caption{Experimental results on varying codebook size $N$, temporal factor $s$, and RQ layers when training RQ-VAE. Both FGD and L1 loss were computed based on RQ-VAE reconstruction results.}
 \label{tab:RQVAE}
\end{table}

\begin{table}
\begin{tabular}
{ 
  | >{\centering\arraybackslash}X
  | >{\centering\arraybackslash}X 
  | >{\centering\arraybackslash}X
  | >{\centering\arraybackslash}X
  | >{\centering\arraybackslash}X
  | >{\centering\arraybackslash}X
  | >{\centering\arraybackslash}X | }
 \hline
 \multicolumn{7}{|c|}{Experiments for Transformer Inference} \\
 \hline 
 Codebook Size $N$ & Temporal Factor $s$ & RQ Layer & FGD ($k=1$) & FGD ($k=5$) & FGD ($k=10$) & FGD ($k=50$)\\
 \hline
 256 & 4 & 2 & 1.642 & 1.669 & 1.620 & 1.604 \\
 256 & 4 & 4 & 1.920 & 1.743 & 1.664 & 1.523 \\
 256 & 2 & 2 & 1.846 & 1.803 & 1.686 & 1.680 \\
 256 & 2 & 4 & 2.036 & 1.873 & 1.733 & 1.605 \\
 \hline
\end{tabular}
 
 \caption{Experimental results on varying top-$k$ parameters during inference. Here the FGD score in each column is computed using different $k$ in top-$k$ sampling.}
  \label{tab:RQTransformer}
\end{table}

\section{Model Architectures}
We listed the detailed model architectures in Table \ref{tbl:component}, \ref{tbl:vqvae}, and \ref{tbl:transformer}. Among them, Table \ref{tbl:component} describes the common building blocks that will be re-used in both models. 

Table \ref{tbl:vqvae} describes the RQ-VAE architecture for learning the discrete gesture tokens. For RQ-VAE with temporal reduction factor $s=4$, the encoder includes stacks of residual connections and 2 downsampling layers (convolution layer with stride of 2). The decoder has a symmetrical architecture with stacks of residual connections and 2 upsampling layers(transposed convolution layer).

Table \ref{tbl:transformer} describes the RQ-Transformer architecture that learns the conditional prior probabilities of gesture tokens. The model mainly consists of 4 parts, the audio encoder, the text encoder, the spatial transformer, and the depth transformer. The audio encoder is a network with 1D convolutions. Text encoder is based on a temporal convolutional network(TCN) with stacks of dilated convolution layers \cite{BaiTCN2018}. Spatial transformer and depth transformer are both stacks of self-attention layers \cite{vaswani2017attention}.

We have included the model source code as part of the supplementary material. The full source code will also be publicly available later via GitHub.




\begin{center}

\begin{longtable}{lll}
\caption[Block Components]{Block Components} \label{Block Components} \\

\hline \multicolumn{1}{c}{\textbf{Components}} & \multicolumn{1}{c}{\textbf{Architecture}}\\ 
\hline 
\endfirsthead

\multicolumn{2}{c}%
{{\bfseries \tablename\ \thetable{} -- continued from previous page}} \\
\hline \multicolumn{1}{c}{\textbf{Components}} &
\multicolumn{1}{c}{\textbf{Architecture}}\\ \hline 
\endhead
            
\hline
\multirow{5}{4em}{ResnetBlock}
& 	(norm1): GroupNorm(32, 64, eps=1e-06) \\
&            	(conv1): Conv1d(64, 64, kernel\textunderscore size=(3,), stride=(1,), padding=(1,)) \\
&            	(norm2): GroupNorm(32, 64, eps=1e-06) \\
&            	(dropout): Dropout(p=0.0) \\
&            	(conv2): Conv1d(64, 64, kernel\textunderscore size=(3,), stride=(1,), padding=(1,)) \\

\hline
\multirow{5}{4em}{ConvAttnBlock}
&            	(norm): GroupNorm(32, 256, eps=1e-06) \\
&            	(q): Conv1d(256, 256, kernel\textunderscore size=(1,), stride=(1,)) \\
&            	(k): Conv1d(256, 256, kernel\textunderscore size=(1,), stride=(1,)) \\
&            	(v): Conv1d(256, 256, kernel\textunderscore size=(1,), stride=(1,)) \\
&            	(proj\textunderscore out): Conv1d(256, 256, kernel\textunderscore size=(1,), stride=(1,)) \\

\hline
\multirow{18}{4em}{TemporalBlock}
&        	(conv1): Conv1d(300,300,kernel\textunderscore size=(2,), stride=(1,), padding=(1,)) \\
&        	(chomp1): Chomp1d() \\
&        	(relu1): ReLU() \\
&        	(dropout1): Dropout(p=0.1, inplace=False) \\
&        	(conv2): Conv1d(300, 300, kernel\textunderscore size=(2,), stride=(1,), padding=(1,)) \\
&        	(chomp2): Chomp1d() \\
&        	(relu2): ReLU() \\
&        	(dropout2): Dropout(p=0.1, inplace=False) \\
&        	(net): Sequential( \\
&          	(0): Conv1d(300, 300, kernel\textunderscore size=(2,), stride=(1,), padding=(1,)) \\
&          	(1): Chomp1d() \\
&          	(2): ReLU() \\
&          	(3): Dropout(p=0.1, inplace=False) \\
&          	(4): Conv1d(300, 300, kernel\textunderscore size=(2,), stride=(1,), padding=(1,)) \\
&          	(5): Chomp1d() \\
&          	(6): ReLU() \\
&          	(7): Dropout(p=0.1, inplace=False) \\
&        	) \\
&        	(relu): ReLU() \\

\hline
\multirow{10}{4em}{SelfAttnBlock}
& (ln1): LayerNorm((256,), eps=1e-05) \\
&        	(ln2): LayerNorm((256,), eps=1e-05) \\
&        	(attn): MultiSelfAttention( \\
&          	(key): Linear(in\textunderscore features=256, out\textunderscore features=256) \\
&          	(query): Linear(in\textunderscore features=256, out\textunderscore features=256) \\
&       	(value): Linear(in\textunderscore features=256, out\textunderscore features=256) \\
&          	(attn\textunderscore drop): Dropout(p=0.1) \\
&          	(resid\textunderscore drop): Dropout(p=0.1) \\
&         	(proj): Linear(in\textunderscore features=256, out\textunderscore features=256) \\
\hline
\hline
\label{tbl:component}
\end{longtable}
\end{center}

\begin{center}
\begin{longtable}{lll}
\caption[RQVAE Architecture]{RQVAE Architecture} \label{RQVAE Architecture} \\

\hline \multicolumn{1}{c}{\textbf{Components}} & \multicolumn{1}{c}{\textbf{Architecture}}\\ 
\hline 
\endfirsthead

\multicolumn{2}{c}%
{{\bfseries \tablename\ \thetable{} -- continued from previous page}} \\
\hline \multicolumn{1}{c}{\textbf{Components}} &
\multicolumn{1}{c}{\textbf{Architecture}}\\ \hline 
\endhead
            
\hline
\multirow{16}{4em}{VAE Encoder}
& (conv\textunderscore in): Conv1d(126, 64, kernel\textunderscore size=(3,), stride=(1,), padding=(1,)) \\ 
& (res\textunderscore block0): ResnetBlock() \\
& (res\textunderscore block1): ResnetBlock() \\
& (down\textunderscore sample0): Downsample(
          	Conv1d(64, 64, kernel\textunderscore size=(3,), stride=(2,))
        	) \\
& (res\textunderscore block2): ResnetBlock() \\
& (res\textunderscore block3): ResnetBlock() \\
& (down\textunderscore sample0): Downsample(
          	Conv1d(128, 128, kernel\textunderscore size=(3,), stride=(2,))
        	) \\
& (res\textunderscore block4): ResnetBlock() \\
& (res\textunderscore block5): ResnetBlock() \\

& (conv\textunderscore attn\textunderscore block0): ConvAttnBlock() \\
& (conv\textunderscore attn\textunderscore block1): ConvAttnBlock() \\
& (res\textunderscore block6): ResnetBlock() \\
& (conv\textunderscore attn\textunderscore block2): ConvAttnBlock() \\
& (res\textunderscore block7): ResnetBlock() \\
& (norm\textunderscore out): GroupNorm(32, 256, eps=1e-06) \\
& (conv\textunderscore out): Conv1d(256, 64, kernel\textunderscore size=(3,), stride=(1,), padding=(1,)) \\ 

\hline
\multirow{4}{4em}{Residual Codebook}
& (0): VQEmbedding(257, 64, padding\textunderscore idx=256) \\ 
& (1): VQEmbedding(257, 64, padding\textunderscore idx=256) \\ 
& (2): VQEmbedding(257, 64, padding\textunderscore idx=256) \\
& (3): VQEmbedding(257, 64, padding\textunderscore idx=256) \\    

\hline
\multirow{12}{4em}{VAE Decoder}
& (conv\textunderscore in): Conv1d(64, 256, kernel\textunderscore size=(3,), stride=(1,), padding=(1,)) \\ 
& (res\textunderscore block0): ResnetBlock() \\
& (conv\textunderscore attn\textunderscore block0): ConvAttnBlock() \\
& (res\textunderscore block1): ResnetBlock() \\
& (res\textunderscore block2): ResnetBlock() \\
& (res\textunderscore block3): ResnetBlock() \\
& (res\textunderscore block4): ResnetBlock() \\
& (res\textunderscore block5): ResnetBlock() \\
& (res\textunderscore block6): ResnetBlock() \\
& (res\textunderscore block7): ResnetBlock() \\
& (up\textunderscore sample0): Upsample(
          	Conv1d(128, 128, kernel\textunderscore size=(3,), stride=(1,))
        	) \\
& (res\textunderscore block8): ResnetBlock() \\
& (res\textunderscore block9): ResnetBlock() \\
& (res\textunderscore block10): ResnetBlock() \\
& (conv\textunderscore attn\textunderscore block1): ConvAttnBlock() \\
& (conv\textunderscore attn\textunderscore block2): ConvAttnBlock() \\
& (conv\textunderscore attn\textunderscore block3): ConvAttnBlock() \\
& (up\textunderscore sample0): Upsample(
          	Conv1d(256, 256, kernel\textunderscore size=(3,), stride=(1,))
        	) \\
& (norm\textunderscore out): GroupNorm(32, 64, eps=1e-06) \\
& (conv\textunderscore out): Conv1d(64, 126, kernel\textunderscore size=(3,), stride=(1,), padding=(1,)) \\ 
\hline
\label{tbl:vqvae}
\end{longtable}
\end{center}

\begin{center}
\begin{longtable}{lll}
\caption[Transformer Architecture]{Transformer Architecture} \label{Transformer Architecture} \\

\hline \multicolumn{1}{c}{\textbf{Components}} & \multicolumn{1}{c}{\textbf{Architecture}}\\ 
\hline 
\endfirsthead

\multicolumn{2}{c}%
{{\bfseries \tablename\ \thetable{} -- continued from previous page}} \\
\hline \multicolumn{1}{c}{\textbf{Components}} &
\multicolumn{1}{c}{\textbf{Architecture}}\\ \hline 
\endhead

\hline
\multirow{13}{4em}{Audio Encoder}
& (Feat Extractor): Sequential( \\
& (0): Conv1d(1, 16, kernel\textunderscore size=(15,), stride=(4,), padding=(1600,)) \\
& (1): BatchNorm1d(16, eps=1e-05, momentum=0.1) \\
& (2): LeakyReLU(negative\textunderscore slope=0.3, inplace=True) \\
& (3): Conv1d(16, 32, kernel\textunderscore size=(15,), stride=(5,)) \\
& (4): BatchNorm1d(32, eps=1e-05, momentum=0.1) \\
& (5): LeakyReLU(negative\textunderscore slope=0.3, inplace=True) \\
& (6): Conv1d(32, 64, kernel\textunderscore size=(15,), stride=(6,), padding=(2,)) \\
& (7): BatchNorm1d(64, eps=1e-05, momentum=0.1) \\
& (8): LeakyReLU(negative\textunderscore slope=0.3, inplace=True) \\
& (9): Conv1d(64, 32, kernel\textunderscore size=(15,), stride=(7,), padding=(2,)) \\
& ) \\
\hline

            \multirow{12}{4em}{Text Encoder}
& (Embedding): Embedding(4040, 300) \\
& (TCN): Sequential( \\
&      	(0): TemporalBlock() \\
&        (1): TemporalBlock() \\
&        (2): TemporalBlock() \\
&        (3): TemporalBlock() \\
&        (4): TemporalBlock() \\
&        (5): TemporalBlock() \\
&        (6): TemporalBlock() \\
&        (7): TemporalBlock() \\
& (decoder): Linear(in\textunderscore features=300, out\textunderscore features=32, bias=True) \\
&  	(drop): Dropout(p=0.1, inplace=False) \\
&	) \\
\hline
 \multirow{10}{4em}{Spacial Transformer}
& (blocks): SelfAttnBlock( \\
& (0): SelfAttnBlock() \\
& (1): SelfAttnBlock() \\
& (2): SelfAttnBlock() \\
& (3): SelfAttnBlock() \\
& (4): SelfAttnBlock() \\
& (5): SelfAttnBlock() \\
& (6): SelfAttnBlock() \\
& (7): SelfAttnBlock() \\
&      	) \\

\hline
 \multirow{6}{4em}{Depth Transformer}
& (blocks): SelfAttnBlock( \\
& (0): SelfAttnBlock() \\
& (1): SelfAttnBlock() \\
& (2): SelfAttnBlock() \\
& (3): SelfAttnBlock() \\
&      	) \\

\hline
 \multirow{6}{4em}{Classifier}
& (classifier) Sequential( \\
&    	(layer\textunderscore norm): LayerNorm((256,), eps=1e-05) \\
&    	(linear): Linear(in\textunderscore features=256, out\textunderscore features=256, bias=True) \\
&    	(logit\textunderscore mask): LogitMask() \\
&  	) \\

\hline
\label{tbl:transformer}
\end{longtable}
\end{center}

{\small
\bibliographystyle{ieee_fullname}
\bibliography{shortstrings,bibliography}
}